\documentclass[lettersize,journal]{IEEEtran}
\usepackage{amsmath,amsfonts}
\usepackage{array}
\usepackage[caption=false,font=normalsize,labelfont=sf,textfont=sf]{subfig}
\usepackage{textcomp}
\usepackage{stfloats}
\usepackage{url}
\usepackage{verbatim}
\usepackage{graphicx}
\usepackage{cite}

\usepackage{multirow}
\usepackage{mathtools}
\usepackage{makecell}
\usepackage{algorithmic}
\usepackage[linesnumbered,boxed,ruled,commentsnumbered]{algorithm2e}
\usepackage{setspace}

\usepackage{amssymb}
\usepackage{fancyhdr}
\usepackage{colortbl}  
\usepackage{xcolor}
\usepackage{array}
\usepackage{booktabs}
\usepackage{amsmath}
\usepackage[switch]{lineno}

\begin{document}

\title{A New Teacher-Reviewer-Student Framework for \\ Semi-supervised 2D Human Pose Estimation}

\author{Wulian~Yun,~Mengshi~Qi,~\IEEEmembership{Member,~IEEE,}~Fei~Peng,~and~Huadong~Ma,~\IEEEmembership{Fellow,~IEEE}
}

\maketitle

\begin{abstract}
Conventional 2D human pose estimation methods typically require extensive labeled annotations, which are both labor-intensive and expensive. In contrast, semi-supervised 2D human pose estimation can alleviate the above problems by leveraging a large amount of unlabeled data along with a small portion of labeled data. 
Existing semi-supervised 2D human pose estimation methods update the network through backpropagation, ignoring crucial historical information from the previous training process. Therefore, we propose a novel semi-supervised 2D human pose estimation method by utilizing a newly designed \emph{Teacher-Reviewer-Student} framework. Specifically, we first mimic the phenomenon that human beings constantly review previous knowledge for consolidation to design our framework, in which the teacher predicts results to guide the student's learning and the reviewer stores important historical parameters to provide additional supervision signals.
Secondly, we introduce a Multi-level Feature Learning strategy, which utilizes the outputs from different stages of the backbone to estimate the heatmap to guide network training, enriching the supervisory information while effectively capturing keypoint relationships.
Finally, we design a data augmentation strategy, \emph{i.e.}, Keypoint-Mix, to perturb pose information by mixing different keypoints, thus enhancing the network's ability to discern keypoints. Extensive experiments on publicly available datasets, demonstrate our method achieves significant improvements compared to the existing methods.   
\end{abstract}


\section{Introduction}
\label{sec:intro}

2D Human pose estimation~(HPE)~\cite{6909610,8100084,10.1007/978-3-030-01231-1_29,8953615,9052469,9710108} aims to infer human pose information from images,~\emph{i.e.}, the positions of keypoints and the connected relationships among them.
As an essential task in the multimedia domain, 
it is widely applied in many downstream tasks,
~\emph{e.g.}, action recognition~\cite{10025821,qi2018stagnet}, person re-identification~\cite{10339887,fu2024mutual}, 3D human pose estimation~\cite{10278485,10374279}, etc. 

Existing 2D HPE methods are primarily categorized into heatmap-based and regression-based.
The heatmap-based methods~\cite{10.1007/978-3-030-01231-1_29,8953615} estimate a likelihood heatmap for each keypoint by predicting confidence at each position. Compared to regression-based methods~\cite{6909610,9710108} that directly predict the keypoint coordinates of images, it can more accurately capture the spatial relationships between keypoints, resulting in better performance. 
Thus, this work mainly focuses on heatmap-based pose estimation.
Nevertheless, these methods require extensive detailed annotations for fully supervised training, limiting their application to areas where annotating large-scale images is costly.

\begin{figure}[t]
    \centering
    \hspace{-3 mm}  
    \includegraphics[scale=.3]{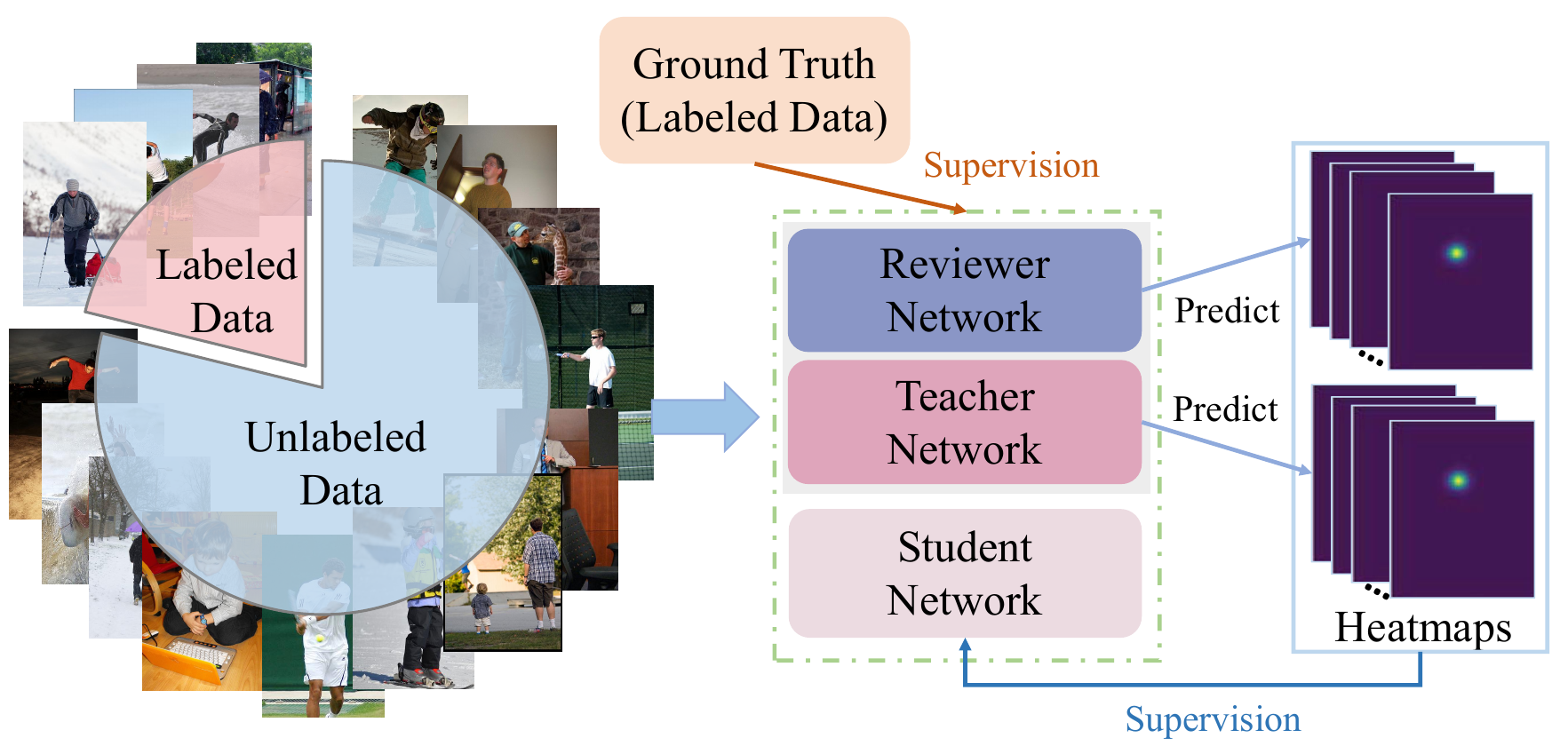}
    \caption{Illustrations of our proposed \emph{Teacher-Reviewer-Student} framework for semi-supervised 2D HPE task. 
    Unlike fully supervised methods that rely solely on labeled data for pose estimation, our semi-supervised method utilizes both labeled and unlabeled data to estimate human pose. Furthermore, we propose the reviewer network based on the teacher-student framework to provide additional supervisory signals.}
    \label{fig:1}
\end{figure}

One promising solution to address this issue is semi-supervised learning, which leverages both abundant unlabeled data and a limited amount of labeled data to improve the model's predictive capacity, as illustrated in Figure~\ref{fig:1}.
Semi-supervised learning is primarily based on the teacher-student framework and can be classified into consistency-based~\cite{10.5555/3294771.3294885,10.5555/3495724.3496249} and pseudo-label based~\cite{NEURIPS2020_06964dce,Liu_2022_CVPR}.
The former aims to ensure that teacher and student networks make similar predictions for the same data under different perturbations, while the latter employs the teacher network to predict pseudo-labels to supervise the student network. Recent semi-supervised 2D HPE studies mainly concentrate on consistency-based semi-supervised learning. For example, 
Xie~\emph{et al.}~\cite{9710942} perform easy-hard augmentation to perturb the data in the same input and then feed it into the teacher-student framework for consistency learning.
In this way, the parameters of teacher and student networks are updated separately.
Huang~\emph{et al.}~\cite{Huang_2023_CVPR} correct pseudo-labels by computing the inconsistency score between the predicted results generated by two teacher networks, thereby providing more accurate results for student network.

However, these methods utilize backpropagation to optimize the network, which usually focuses only on the parameter updates at the current moment and ignores historical variation of model parameters during training, thus failing to retain essential historical information~\cite{grill2020bootstrap,he2020momentum,10.5555/3294771.3294885,yuan2023semi}. 
In contrast, exponential moving average (EMA)~\cite{10.5555/3294771.3294885} incorporates historical parameter information during training by employing a weighted average of historical model parameters. Therefore, it is worth exploring how to leverage the advantages of EMA to enhance the performance of semi-supervised 2D HPE.

In this study, we introduce a newly designed \emph{Teacher-Reviewer-Student} framework for semi-supervised 2D HPE, which is inspired by the phenomenon that people gradually forget past knowledge as they continue to receive new information, and so they need to consolidate what they have previously learned through review. Especially, we integrate two reviewer networks into the existing teacher-student framework, in which the teacher network and student network update parameters separately by alternating roles with each other, and the reviewer networks retain historical parameter information of both the teacher and student networks during the training. 
More precisely, the parameters of the reviewer networks are updated by the teacher and student networks via EMA after each training step.
This enables the reviewer networks to promptly consolidate key training information by aggregating the weights learned by both the teacher and student networks. 
 
Furthermore, how to uncover extra supervisory information from unlabeled data is crucial in semi-supervised learning. 
While we have explored training manner, we aim to further uncover potential supervisory information from the data and its features to enhance the model's learning ability and improve pose estimation performance with limited labeled data.
From the feature perspective, current semi-supervised 2D HPE methods exploit the final output generated by the backbone to estimate the heatmap.
However, this manner focuses primarily on the semantic information in deep features while ignoring the spatial information present in features from other stages. 
Therefore, we propose a Multi-level Feature Learning strategy, which involves upsampling the outputs of the multiple stages from the backbone to generate heatmaps. This strategy integrates information across stages to learn keypoint relationships, thereby enhancing estimation accuracy. 

In terms of data, considering that data augmentation is frequently employed in semi-supervised training to generate hard samples, we design a new data augmentation strategy, named Keypoint-Mix, which scrambles the keypoint information by mixing image patches around different keypoints, and then covers the blended generated image patch back to the original region. This strategy perturbs the pose information while preserving crucial pose details, thereby enhancing the network's capability to distinguish keypoints.

The contributions can be summarized as follows:

\par\textbf{(1)}
We propose a novel Teacher-Reviewer-Student  framework for semi-supervised 2D HPE, where reviewer networks store crucial training information of both teacher and student networks to provide more supervision.

\par\textbf{(2)}
We design a Multi-level Feature Learning strategy to enrich the supervisory signals, which utilize different levels of features to learn the relationships between keypoints.

\par\textbf{(3)}
We introduce the Keypoint-Mix to perturb pose information, enabling the network to better discern keypoints.

\par\textbf{(4)}
Extensive experiments demonstrate that our proposed method obtains substantial improvements and achieves state-of-the-art performance.

The rest of this paper is organized as follows. Section~\uppercase\expandafter{\romannumeral2} summarizes recent progress in 2D human pose estimation and semi-supervised learning. Then, section~\uppercase\expandafter{\romannumeral3} expresses the preliminaries of our method. Second, section~\uppercase\expandafter{\romannumeral4} presents the proposed semi-supervised 2D human pose estimation method in detail. Afterward, experimental results and discussions are reported in section~\uppercase\expandafter{\romannumeral5}. Finally, section~\uppercase\expandafter{\romannumeral6} draws the conclusion.

\section{Related Work}
This section summarizes recent advances in 2D human pose estimation and semi-supervised Learning domains.
\label{sec:rel}

\noindent
\textbf{2D Human Pose Estimation.}
Current 2D HPE methods ~\cite{6909610,9710108,10.1007/978-3-031-20068-7_6,tian2019directpose,10.1007/978-3-030-01231-1_29,8953615,9157744,Ye_2023_CVPR,newell2016stacked,ke2018multi,sun2019deep,huang2020devil,zhang2019human,wang2023decenternet} can be divided into two classes: regression-based and heatmap-based. 
Regression-based methods~\cite{6909610,9710108,10.1007/978-3-031-20068-7_6,tian2019directpose} can directly predict the keypoint coordinates of the input data. 
Direct regress joint coordinates is first proposed by Toshev~\emph{et al.}~\cite{6909610}.
DirectPose~\cite{tian2019directpose} performs multi-person human pose estimation within a one-stage object detection framework that directly regresses joint coordinates rather than bounding boxes.
SimCC~\cite{10.1007/978-3-031-20068-7_6} represents pose estimation as two classification tasks in horizontal and vertical coordinates. 
RLE~\cite{9710108} captures the underlying distribution by normalizing the flow and then utilizes residual log-likelihood estimation to learn the variation between the original and underlying distributions.
Heatmap-based methods~\cite{10.1007/978-3-030-01231-1_29,8953615,9157744} predict scores for each location to represent the confidence that the location belongs to a keypoint, thus generating a likelihood heatmap for each keypoint. 
For example, SimpleBaseline~\cite{10.1007/978-3-030-01231-1_29} 
generate heatmaps from low-resolution images by introducing deconvolutional layers to scale features to the original image size.
HRNet~\cite{8953615} parallelizes multiple resolution branches while interacting information between different branches to improve the performance. 
Heatmap-based methods can explicitly learn more spatial information by generating probability than regression-based methods, so we primarily focus on such methods. 
However, above methods typically undergo training in a fully supervised way, and the process of data labeling is time-consuming and laborious. In contrast, our method aims to enhance 2D HPE performance with only a limited number of labeled data.

\noindent
\textbf{Semi-supervised Learning.}
Semi-supervised learning~\cite{yun2024weakly,qi2019ke,yun2025semi,10.5555/3294771.3294885,10.5555/3495724.3496249,9156610,Liu_2022_CVPR} plays an important role in tackling computer vision problems, as it effectively leverages both a large amount of unlabeled data and a small portion of labeled data. 
Existing semi-supervised learning contains two dominant approaches: consistency-based~\cite{10.5555/3294771.3294885,10.5555/3495724.3496249} and pseudo-label based~\cite{9156610,Liu_2022_CVPR}.
Consistency-based methods leverage regularization loss to ensure that teacher and student networks with different perturbation inputs can have consistent predictions for the same data.
Mean Teacher~\cite{10.5555/3294771.3294885} employs the exponential moving average (EMA)~\cite{10.5555/3294771.3294885} to update the parameters of the student network to the teacher network, where the teacher and student networks handle strong and weak perturbation data, respectively.
Pseudo-label based methods employ the teacher network to predict pseudo-label for unlabeled data, providing supervision for the student.
FixMatch~\cite{NEURIPS2020_06964dce} generates pseudo-labels from weak augmentation unlabeled samples and subsequently utilizes them to supervise the predictions of strong augmentation samples. 

Semi-supervised 2D HPE \cite{9710942,Huang_2023_CVPR} aims to maximize the model's pose estimation capability by effectively leveraging both labeled and unlabeled data. Xie~\emph{et al.}~\cite{9710942} conduct easy-hard augmentation to the same input, 
which is then fed into a teacher-student framework for consistency learning between the predictions for easy and hard augmented data. 
During this process, the roles of teacher and student are alternated to update the parameters.
Huang~\emph{et al.}~\cite{Huang_2023_CVPR} observed that noisy pseudo-labels affect the training process.
Therefore, they employ two teacher networks to generate pseudo-labels for supervising the student network's training, in which the outliers are removed by calculating the inconsistency score between the two teachers' pseudo-labels.	 
Unlike the above methods, we recognize the significance of historical parameter information. Therefore, we introduce the Teacher-Reviewer-Student framework to effectively leverage unlabeled data, in which the teacher network predicts results to guide the training of the student network, and the reviewer network enhances supervision by storing essential historical parameter training information.

\begin{figure*}[ht]
\begin{center}
\setlength{\fboxrule}{0pt}
\setlength{\fboxsep}{0cm}
\hspace{-2.5mm}
\includegraphics[width=0.87\linewidth]{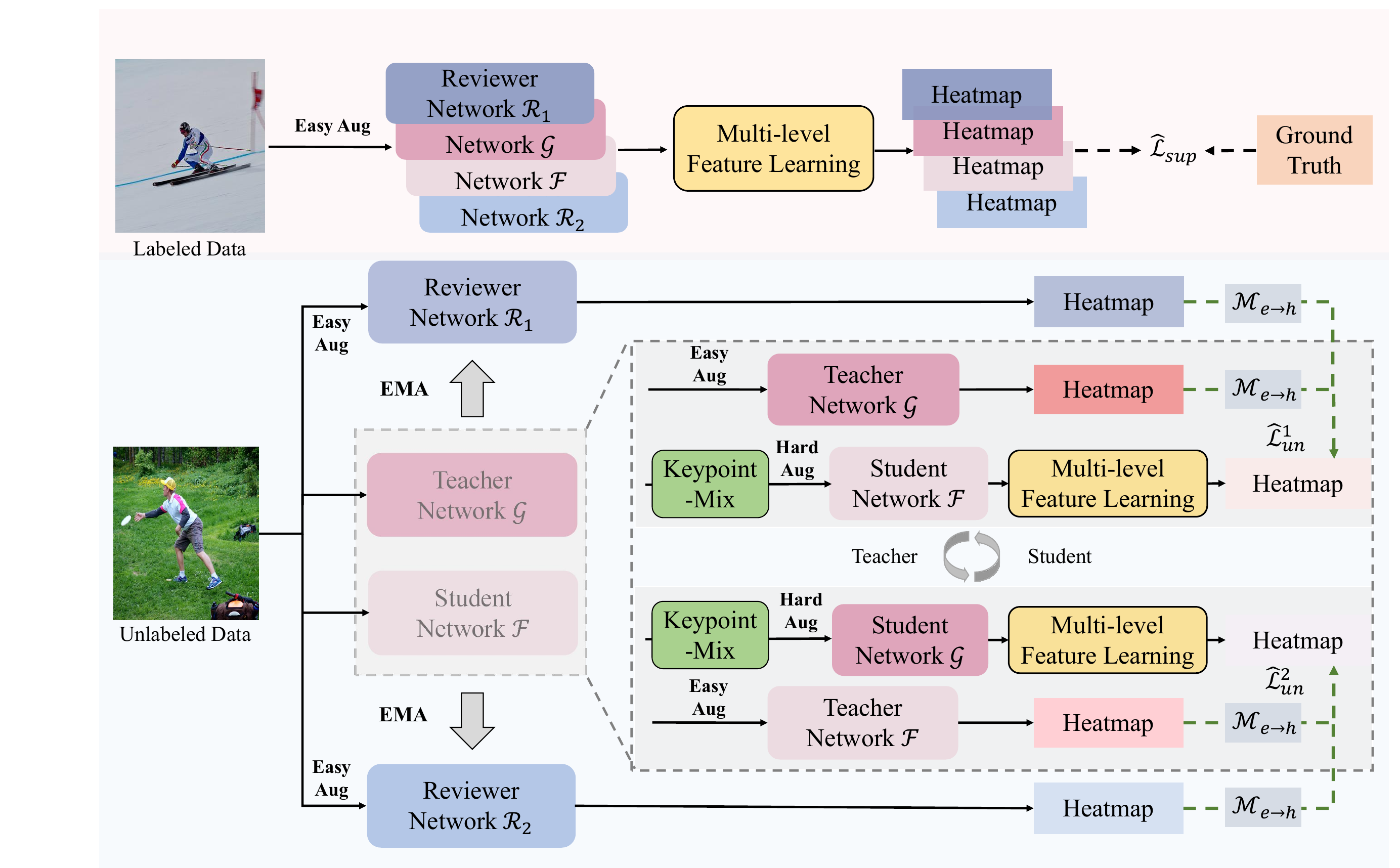}
\end{center}
\caption{Overview of our framework. Our method comprises network $\mathcal{G}$, network $\mathcal{F}$ and reviewer networks $\mathcal{R}_1$, $\mathcal{R}_2$. Network $\mathcal{G}$ and network $\mathcal{F}$ both take turns playing the roles of teacher and student. The teacher network generates predicted results for unlabeled data to guide the training of the student network. Reviewer networks retain crucial information from network $\mathcal{G}$ and network $\mathcal{F}$ during the training while providing additional supervision, which parameters are updated from network $\mathcal{G}$ and network $\mathcal{F}$ via EMA. Multi-level Feature Learning indicates upsampling the outputs of the multiple stages of the backbone to estimate the heatmap.  Keypoint-Mix is a data augmentation strategy.  ${\mathcal{M}_{e \rightarrow h}}$ denotes the mapping of the predicted results of easy augmented data and hard augmented data to the same coordinate space.}
  \label{fig:2}
\end{figure*}

\section{Preliminaries}
In this section, we will describe the problem definition of the semi-supervised 2D HPE task and provide details of the teacher-student framework.

\textbf{Problem definition.} 
In semi-supervised 2D HPE, given a training set $D$ comprising a labeled subset $D^l=\{(I_i^l, S_i^l ) \}_{i=1}^N$ and a unlabeled subset $D^u= \{I_j^u\}_{j=1}^M$. Specifically, $I_i^l$ and $I_j^u$ denote labeled and unlabeled images, $S_i^l$ represents the ground truth of $I_j^l$, $N$ and $M$ refer to the number of labeled and unlabeled data. In practice, where $N \ll M$.
Semi-supervised 2D HPE aims to simultaneously leverage the limited labeled data  $D^l$ as well as unlabeled data $D^u$ for training, and then perform pose estimation for each test image to get the positions of different keypoints.

\noindent\textbf{Teacher-student framework.} 
We adopt the traditional teacher-student framework as a basis for our semi-supervised method. Specifically, this framework contains the teacher network and the student network, where the parameters $\theta_t$ of the teacher network are updated from the student network’s parameters $\theta_s$ by EMA~\cite{10.5555/3294771.3294885}: 
\begin{equation}
\label{eq:1}
\theta_t=\eta \theta_t+(1-\eta) \theta_s
\end{equation}
where $\eta \in(0,1)$ indicates the momentum. During the training phase, labeled data is fed into the student network, and the difference between the predicted results of the student network and the ground truth is calculated to optimize the student network.
For unlabeled data, different data augmentations are first performed and then the augmented data are sent to the teacher network and student network, respectively. Then, the teacher network generates pseudo-labels to guide the training of the student network.

\section{Proposed Method}
\label{sec:method}
In this section, we will begin by presenting a pipeline of our proposed method in~\uppercase\expandafter{\romannumeral4}-A. Next, we will describe the different components of our method in detail.
First, we describe the Teacher-Reviewer-Student framework in~\uppercase\expandafter{\romannumeral4}-B. Then, the Multi-level Feature Learning is presented in~\uppercase\expandafter{\romannumeral4}-C, which is used to enrich the
supervisory signal by leveraging different levels of features. 
Second, we describe the data augmentation strategy Keypoint-Mix in ~\uppercase\expandafter{\romannumeral4}-D, which enhances the network’s
capability to distinguish keypoint by perturbing the pose information.
Finally, we introduce the training details of optimizing the  Teacher-Reviewer-Student framework in section~\uppercase\expandafter{\romannumeral4}-E.
The descriptions of all symbols used in our method are shown in Table~\ref{tab:1}.

\subsection{Pipeline}
The overview of our method is shown in Figure~\ref{fig:2}. Our training process contains two main parts:
1) \textbf{For labeled data:} We separately input the labeled data $I_i^l$ into network $\mathcal{G}$, network $\mathcal{F}$ and reviewer networks $\mathcal{R}_1$, $\mathcal{R}_2$ for training and update their parameters accordingly.
2) \textbf{For unlabeled data:} 
During network $\mathcal{G}$ acts as the teacher, network $\mathcal{F}$ plays the role of the student. Specifically, teacher network $\mathcal{G}$ and reviewer network  $\mathcal{R}_1$ generate predicted results to guide student's training. Hard data augmentation is performed to the input data of student network $\mathcal{F}$ and easy data augmentation is applied to the input data of teacher network $\mathcal{G}$ and reviewer network $\mathcal{R}_1$. The student network $\mathcal{F}$ updates its parameters. Conversely, when  network $\mathcal{G}$ takes on the role of the student, network $\mathcal{F}$ acts as the teacher. Predicted results are generated by teacher  network $\mathcal{F}$ and reviewer network  $\mathcal{R}_2$ to guide the student's training. The student network
$\mathcal{G}$ updates its parameters.
Meanwhile, the parameters of the reviewer networks $\mathcal{R}_1$ and $\mathcal{R}_2$ are updated by network $\mathcal{G}$ and network $\mathcal{F}$ through EMA, respectively.
During the training process, we introduce a Multi-level Feature Learning strategy to enrich the supervisory signals by utilizing the output of different stages of the backbone. 
In addition, we design the Keypoint-Mix to provide additional challenging samples by blending information from diverse keypoint positions.

\begin{table}[ht]
\centering
\caption{The list of used symbols and their descriptions in our method.}
\renewcommand{\arraystretch}{1.1}
\setlength{\tabcolsep}{0.9mm}{
\label{tab:1}
\begin{tabular}{c|c}
\specialrule{0.1em}{0.5pt}{0.5pt}
\hline
Symbol & Description \\ \hline
$D^l$   &  Labeled set           \\  
$D^u$  &   Unlabeled set  \\  
$I^l_i$    &  Labeled images       \\  
$I^u_j$      &   Unlabeled images  \\ 
$S_i^l$  & The ground truth of labeled images  \\  
$N$, $M$  & The number of labeled and unlabeled images  \\      $\mathcal{G}$, $\mathcal{F}$ & Teacher network and student network  \\  
$\mathcal{R}_1$, $\mathcal{R}_2$  & 
 Reviewer networks  \\  
$\hat{S}_i^{l^\mathcal{G}_{z}}$, $\hat{S}_i^{l^\mathcal{G}_{p}}$ 
& Predicted results of network $\mathcal{G}$ on labeled data
\\  
$\hat{S}_i^{l^\mathcal{F}_{z}}$, $\hat{S}_i^{l^\mathcal{F}_{p}}$ 
& Predicted results of network $\mathcal{F}$ on labeled data
\\  
 $\hat{S}_i^{l^{\mathcal{R}_1}_{z}}$, $\hat{S}_i^{l^{\mathcal{R}_1}_{p}}$ 
& Predicted results of network $\mathcal{R}_1$  on labeled data
\\  
 $\hat{S}_i^{l^{\mathcal{R}_2}_{z}}$, $\hat{S}_i^{l^{\mathcal{R}_2}_{p}}$ 
& Predicted results of network $\mathcal{R}_2$ on labeled data
\\  
$\tilde{S}_j^{u^{\mathcal{G}}}$,  $\tilde{S}_j^{u^{\mathcal{R}_1}}$
& \makecell[c]{Predicted results of teacher network $\mathcal{G}$ \\ and reviewer network $\mathcal{R}_1$ on unlabeled data}
\\   
$\bar{S}_j^{u^\mathcal{F}_z}$,  $\bar{S}_j^{u^\mathcal{F}_p} $
&  \makecell[c]{Predicted results of student network\\ $\mathcal{F}$ on unlabeled data}
\\   
$\tilde{S}_j^{u^{\mathcal{F}}}$,  $\tilde{S}_j^{u^{\mathcal{R}_2}}$
&   \makecell[c]{Predicted results of teacher network $\mathcal{F}$ \\ and reviewer network $\mathcal{R}_2$ on unlabeled data}
\\  
 $\bar{S}_j^{u^\mathcal{G}_z}$ , $\bar{S}_j^{u^\mathcal{G}_p} $ 
 &    \makecell[c]{Predicted results of student network \\ $\mathcal{G}$ on unlabeled data} \\  
$\theta_{\mathcal{G}}$, $\theta_{\mathcal{F}}$  & The parameters of network $\mathcal{G}$ and network$\mathcal{F}$ \\ 
$\theta_{\mathcal{R}_1}$, $\theta_{\mathcal{R}_2}$  & The parameters of network $\mathcal{R}_1$ and network $\mathcal{R}_2$ \\  
$\mathcal{L}$     &       The loss function of our method      \\  \hline
\specialrule{0.1em}{1pt}{1pt}
\end{tabular}
}
\end{table}

\subsection{Teacher-Reviewer-Student Framework}
Different from traditional teacher-student framework, where the roles of teacher and student are fixed and the teacher’s parameters are updated by the student. 
Our method contains a network $\mathcal{G}$, a network $\mathcal{F}$ and two reviewer networks $\mathcal{R}_1$, $\mathcal{R}_2$. Specifically, network $\mathcal{G}$ and network $\mathcal{F}$ alternate between acting as teacher and student by inputting different augmented data. The parameters of both networks remain independent.
Reviewer networks $\mathcal{R}_1$, $\mathcal{R}_2$ are used to record the crucial parameters information of network $\mathcal{G}$ and network $\mathcal{F}$ during the training process and then provide additional information to guide the training of the student network. The parameters of reviewer networks $\mathcal{R}_1$, $\mathcal{R}_2$ are updated by the network $\mathcal{G}$ and  network $\mathcal{F}$, respectively.
In the training of a batch, both labeled and unlabeled data are included simultaneously.

\noindent
\textbf{Labeled data.}
% \noindent
% \textbf{Learning with ground truth.}
For each labeled data $I_i^l$, it needs to learn with the ground truth. We first feed it into network $\mathcal{G}$, network $\mathcal{F}$, reviewer networks $\mathcal{R}_1$ and $\mathcal{R}_2$ to obtain predicted results $\hat{S}_i^{l^\mathcal{G}}$, $\hat{S}_i^{l^\mathcal{F}}$ $\hat{S}_i^{l^{\mathcal{R}_1}}$ and   $\hat{S}_i^{l^{\mathcal{R}_2}}$. 
It can be defined as:
\begin{equation}
\label{eq:2}
\begin{aligned}
  \hat{S}_i^{l^\mathcal{W}}= \mathcal{W}(I_i^l),\quad \mathcal{W}\in\{\mathcal{G},\mathcal{F},\mathcal{R}_1,\mathcal{R}_2\}
\end{aligned}
\end{equation}
where $\mathcal{G} (\cdot)$,  $\mathcal {F} (\cdot)$, ${\mathcal{R}_1} (\cdot)$ and ${\mathcal{R}_2} (\cdot)$ denote the networks $\mathcal{G}$, $\mathcal {F} $, $\mathcal{R}_1$ and  $\mathcal{R}_2$, respectively.

Then, we utilize the Mean Squared Error (MSE) to calculate the difference between the predicted results and the ground truth, which is used to train and update the parameters $\theta_\mathcal{G}$, $\theta_\mathcal{F}$, $\theta_{\mathcal{R}_1}$, $\theta_{\mathcal{R}_2}$ of networks $\mathcal{G}$, $\mathcal{F}$, $\mathcal{R}_1$ $\mathcal{R}_2$. The loss function can be defined as
\begin{equation}
\label{eq:3}
\mathcal{L}_{sup}=\sum_{i\in N}\sum_{\mathcal{W}\in\{\mathcal{G},\mathcal{F},\mathcal{R}_1,\mathcal{R}_2\}}(\hat{S}_i^{l^\mathcal{W}}-S_i^l)^2
\end{equation}
where $S_i^l$ denotes ground truth of $I_i^l$.

\noindent
\textbf{Unlabeled data.}
For each unlabeled data $I_j^u$, we first perform easy data augmentation on it. Then, we pass it through network $\mathcal{G}$, network $\mathcal{F}$, reviewer networks $\mathcal{R}_1$ and $\mathcal{R}_2$ to obtain different predicted results $\tilde{S}_j^{u^\mathcal{G}}$, $\tilde{S}_j^{u^\mathcal{F}}$, $\tilde{S}_j^{u^{\mathcal{R}_1}}$ and $\tilde{S}_j^{u^{\mathcal{R}_2}}$, respectively. The process can be formulated as: 
\begin{equation}
\label{eq:4}
\begin{aligned}
  \tilde{S}_j^{u^\mathcal{W}}= \mathcal{W}(I_j^u),\quad \mathcal{W}\in\{\mathcal{G},\mathcal{F},\mathcal{R}_1,\mathcal{R}_2\}
\end{aligned}
\end{equation}

When network $\mathcal{F}$ is the student, network $\mathcal{G}$ plays the role of the teacher, while reviewer network $\mathcal{R}_1$ provides additional supervision for the student network. For each unlabeled data $I_j^u$, we first perform hard data augmentation and then seed it into 
student network $\mathcal{F}$ to obtain prediction result $\bar{S}_j^{u^\mathcal{F}}$. 
Subsequently, the predicted results $\tilde{S}_j^{u^{\mathcal{G}}}$ and $\tilde{S}_j^{u^{\mathcal{R}_1}}$ generated by teacher network $\mathcal{G}$ and reviewer network $\mathcal{R}_1$, respectively, are used to calculate consistency with the prediction result $\bar{S}_j^{u^\mathcal{F}}$ from the student network.
The process can be formulated as follows:
\begin{equation}
\label{eq:5}
\mathcal{L}_{un}^1=\sum_{j\in M}\sum_{\mathcal{W}\in\{\mathcal{G},\mathcal{R}_1\}}(\bar{S}_j^{u^\mathcal{F}}-\mathcal{M}_{e\to h}(\tilde{S}_j^{u^\mathcal{W}}))^2
\end{equation}
where ${\mathcal{M}_{e \rightarrow h}}$ denotes mapping $\tilde{S}_j^{u^{\mathcal{G}}}, \tilde{S}_j^{u^{\mathcal{R}_1}}$ and $\bar{S}_j^{u^\mathcal{F}}$ to the same coordinate space.  During the above process, the parameters of the student network $\mathcal{F}$ are updated. 
In contrast, when network $\mathcal{G}$ plays the role of the student, network $\mathcal{F}$ as the teacher, while reviewer network $\mathcal{R}_2$ provides additional supervision for the student network: 
\begin{equation}
\label{eq:6}
\mathcal{L}_{un}^2=\sum_{j\in M}\sum_{\mathcal{W}\in\{\mathcal{F},\mathcal{R}_2\}}(\bar{S}_j^{u^\mathcal{G}}-\mathcal{M}_{e\to h}(\tilde{S}_j^{u^\mathcal{W}}))^2
\end{equation}
where $\bar{S}_j^{u^\mathcal{G}}$ denotes the predicted result from the student network $\mathcal{G}$, $\tilde{S}_j^{u^{\mathcal{F}}}, \tilde{S}_j^{u^{\mathcal{R}_2}}$ are the predicted results of teacher network $\mathcal{F}$ and reviewer network $\mathcal{R}_2$.   
The parameters of the student network $\mathcal{G}$ can be updated during the above process. The parameters $\theta_{\mathcal{R}_1}$ of reviewer network $\mathcal{R}_1$ and the parameters $\theta_{\mathcal{R}_2}$ of reviewer network $\mathcal{R}_2$ are updated from parameters $\theta_\mathcal{G}$ of  network $\mathcal{G}$ and parameters $\theta_\mathcal{F}$ of network $\mathcal{F}$ via EMA, respectively. The process can be formulated as: 
\begin{equation}
\label{eq:7}
\begin{aligned}
\theta_{\mathcal{R}_1} & =\alpha \theta_{\mathcal{R}_1} + (1-\alpha) \theta_\mathcal{G} , \\
\theta_{\mathcal{R}_2} & =\beta \theta_{\mathcal{R}_2} + (1-\beta) \theta_\mathcal{F} ,
\end{aligned}
\end{equation}
where  $\alpha \in(0,1)$ and $\beta\in(0,1)$ indicate the momentum. With this updated strategy, the reviewer networks $\mathcal{R}_1$ and $\mathcal{R}_2$ can immediately aggregate previously the weights learned by network $\mathcal{G}$ and network $\mathcal{F}$ after each training step.

\subsection{Multi-level Feature Learning}
Heatmap-based methods~\cite{9710942,Huang_2023_CVPR} usually exploit the last output of the backbone for upsampling to estimate the heatmap. However, these methods primarily emphasize the semantic information of the deep feature while neglecting the spatial information present in features from other stages. Hence, we design a Multi-level Feature Learning strategy to upsample the outputs of the multiple stages of the backbone to estimate the heatmap for supervising training. 
This strategy fully utilizes different level features and offers extra supervision signals to guide training.

Since the optimal results are achieved by using the output features from the last two stages of the backbone network, we use this process as an example to demonstrate how our strategy uncovers additional supervision information. 
Specifically, we designate the output features of the last stage of the backbone as $F_z$ and the output features of the penultimate stage as $F_p$. Subsequently, $F_z$ and $F_p$ are upsampled to the same size to obtain features $\hat{F}_z$ and $\hat{F}_p$. These features are then used to estimate the heatmap and obtain prediction results. Specifically, we fed each labeled data $I_i^l$ into different networks $\mathcal{G}$, $\mathcal{F} $, $\mathcal{R}_1$, $\mathcal{R}_2$ to predict results:
\begin{equation}
\label{eq:8}
\hat{S}_i^{l_z^\mathcal{W}}, \hat{S}_i^{l_p^\mathcal{W}}=\mathcal{W}\left(I_i^l\right), \quad \mathcal{W} \in\left\{\mathcal{G}, \mathcal{F}, \mathcal{R}_1, \mathcal{R}_2\right\}
\end{equation}

Then, we exploit MSE to calculate the difference between the predicted results from different networks and the ground truth, respectively.
The loss function can be as follows:
\begin{equation}
\label{eq:9}
\hat{\mathcal{L}}_{sup}=\sum_{i \in N} \sum_{\mathcal{W} \in\left\{\mathcal{G}, \mathcal{F}, \mathcal{R}_1, \mathcal{R}_2\right\}} \sum_{\mathcal{V} \in\{z, p\}}(\hat{S}_i^{l_\mathcal{V}^\mathcal{W}}-S_i^l)^2
\end{equation}
where $S_i^l$ denotes ground truth of $I_i^l$.

For each unlabeled data $I_j^u$, we utilize predicted results generated by the teacher network and the reviewer network to supervise the student ones.
When network $\mathcal{G}$ acts as the teacher, the consistency loss function is as follows:
\begin{equation}
\label{eq:10}
\hat{\mathcal{L}}_{un}^1=\sum_{j \in M} \sum_{\mathcal{W} \in\left\{\mathcal{G}, \mathcal{R}_1\right\}} \sum_{\mathcal{V} \in\{z, p\}}(\bar{S}_j^{u_\mathcal{V}^\mathcal{F}}-\mathcal{M}_{e \rightarrow h}(\tilde{S}_j^{u^\mathcal{W}}))^2
\end{equation}
where $\bar{S}_j^{u^\mathcal{F}_z}$ and $\bar{S}_j^{u^\mathcal{F}_p} $ denote the predicted results of student network $\mathcal{F}$, 
 $\tilde{S}_j^{u^{\mathcal{G}}}$ and $\tilde{S}_j^{u^{\mathcal{R}_1}}$ indicate the predicted results of teacher network $\mathcal{G}$ and reviewer network $\mathcal{R}_1$, respectively. 
 When network $\mathcal{F}$ serves as the teacher, we calculate the consistency between the predictions of teacher network $\mathcal{F}$ and student network $\mathcal{G}$, as well as between reviewer network $\mathcal{R}_2$ and student network $\mathcal{G}$ separately. 
 The loss function is defined as $\hat{\mathcal{L}}_{un}^2$:
\begin{equation}
\label{eq:11}
\hat{\mathcal{L}}_{un}^2=\sum_{j \in M} \sum_{\mathcal{W} \in\left\{\mathcal{F}, \mathcal{R}_2\right\}} \sum_{\mathcal{V} \in\{z, p\}}(\bar{S}_j^{u_\mathcal{V}^\mathcal{G}}-\mathcal{M}_{e \rightarrow h}(\tilde{S}_j^{u^\mathcal{W}}))^2
\end{equation}

\begin{figure}[t]
\begin{center}
\setlength{\fboxrule}{0pt}
\setlength{\fboxsep}{0cm}
\includegraphics[width=1.0\linewidth]{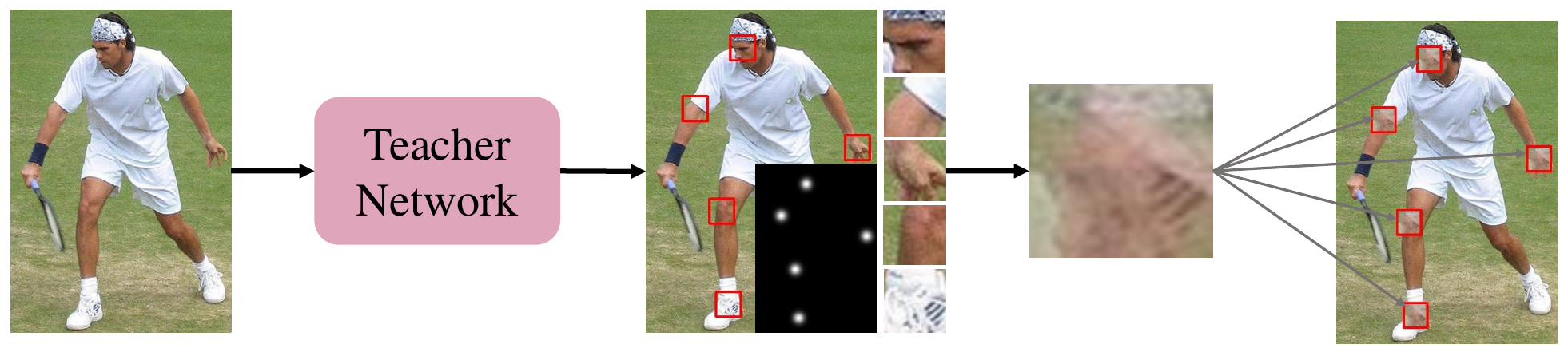}
\end{center}
\caption{Illustration of data augmentation strategy Keypoint-Mix. 
Unlabeled data is fed into the teacher network to generate keypoint predictions, which are then randomly sampled, with image patches extracted from the surrounding regions. Afterward, these image patches are mixed to obtain a blended patch to cover back the original regions. }
\label{fig:3}
\end{figure}

\subsection{Data Augmentation}
Data augmentation can provide additional challenging samples for model training, prompting the model to learn robust feature representations and enhancing its generalization capabilities.
In the 2D HPE task, accurately locating keypoints and learning the relationships between them are crucial. Therefore, we design a data augmentation strategy called Keypoint-Mix, which aims at blending information from different keypoint positions, making it challenging for the model to distinguish which specific category the current keypoint belongs to. This perturbs the pose information while preserving important pose details, thereby increasing the network’s ability to discern keypoints.

Specifically, as shown in Figure~\ref{fig:3}, we first input the image into the teacher network to estimate heatmeap, \emph{i.e.}, the coordinates $\{\{x_1, y_1\},...,\{x_n, y_n\}\}$ of each keypoint. Then we randomly sample $K$ keypoints among these keypoints and extract image patches $\{p_1,...,p_k\}$ from their surrounding regions. 
We then mix these patches using averaging to obtain the blended region patches, and then cover the blended ones back to the original regions of $K$ keypoints. 
Finally, we can generate a hard augmentation sample for each data, which is subsequently fed to the student network for learning.

\subsection{Training Loss}
The final training loss of our method contains both supervised loss $\hat{\mathcal{L}}_{sup}$ and unsupervised loss $\hat{\mathcal{L}}_{un}^1$ and $\hat{\mathcal{L}}_{un}^2$. The total loss function ${\mathcal{L}}$  can be defined as:
\begin{equation}
\label{eq:12}
{\mathcal{L}} =\lambda \cdot \hat{\mathcal{L}}_{sup}+ \hat{\mathcal{L}}_{un}^1 + \hat{\mathcal{L}}_{un}^2 
\end{equation}
where $\lambda$ denotes trade-off factors. The complete training process of our method is described in Algorithm~\ref{alg:1}.

\begin{algorithm}[h]
\setstretch{1}
\caption{The training process of our method.}
\label{alg:1}
\LinesNumbered  
\KwIn{Labeled data ${(I_i^l, S_i^l)}\in D^l$, unlabeled data $I_j^u \in D^u$,  maximum iteration $E$.}
\KwOut{Model $\theta_\mathcal{G}$ of network $\mathcal{G}$, model $\theta_ \mathcal{F}$ of network $\mathcal{F}$.}
\For{$e < E$}{
    ~Sample a batch of labeled data $\{(I_i^l, S_i^l)\}_{i=1}^b$ and unlabeled data $\{I_j^u\}_{j=1}^c$ from $D^l$ and $D^u$\;
    \For {each \text{labeled data} $I_i^l$}{
        ~Predict results $\hat{S}_i^{l^\mathcal{G}_{z}}$, $\hat{S}_i^{l^\mathcal{G}_{p}}$, $\hat{S}_i^{l^\mathcal{F}_{z}}$,  $\hat{S}_i^{l^\mathcal{F}_{p}}$, $\hat{S}_i^{l^{\mathcal{R}_1}_{z}}$, $\hat{S}_i^{l^{\mathcal{R}_1}_{p}}$,  $\hat{S}_i^{l^{\mathcal{R}_2}_{z}}$, $\hat{S}_i^{l^{\mathcal{R}_2}_{p}}$ by teacher network $\mathcal{G}$, student network $\mathcal{F}$, reviewer networks $\mathcal{R}_1$, $\mathcal{R}_2$\;
        Calculate $\hat{\mathcal{L}}_{sup}$ by Eq.\eqref{eq:9}\;
    }
    \For {\text{each unlabeled data} $I_j^u$}{
        ~Predict results $\tilde{S}_j^{u^{\mathcal{G}}}$,  $\tilde{S}_j^{u^{\mathcal{R}_1}}$ by teacher network $\mathcal{G}$, reviewer network $\mathcal{R}_1$, and predict results $\bar{S}_j^{u^\mathcal{F}_z}$,  $\bar{S}_j^{u^\mathcal{F}_p} $ by student network $\mathcal{F}$\;
        Calculate $\hat{\mathcal{L}}_{un}^1$ with $( \bar{S}_j^{u^\mathcal{F}_z}, \bar{S}_j^{u^\mathcal{F}_p}, \tilde{S}_j^{u^{\mathcal{G}}},  \tilde{S}_j^{u^{\mathcal{R}_1}} )$ by Eq.\eqref{eq:10}\;
        Predict results $\tilde{S}_j^{u^{\mathcal{F}}}$,  $\tilde{S}_j^{u^{\mathcal{R}_2}}$ by teacher network $\mathcal{F}$, reviewer network $\mathcal{R}_2$ and predict results $\bar{S}_j^{u^\mathcal{G}_z}$, $\bar{S}_j^{u^\mathcal{G}_p}$ by student network $\mathcal{G}$\;
        Calculate $\hat{\mathcal{L}}_{un}^2$ with $(\bar{S}_j^{u^\mathcal{G}_z}, \bar{S}_j^{u^\mathcal{G}_p}, \tilde{S}_j^{u^{\mathcal{F}}},\tilde{S}_j^{u^{\mathcal{R}_2}})$ by Eq.\eqref{eq:11}\;  
    }
    ~Update $\theta_\mathcal{G}$, $\theta_\mathcal{F}$,  $\theta_{\mathcal{R}_1}$ and $\theta_{\mathcal{R}_2}$ by $ {\mathcal{L}}$ in Eq.\eqref{eq:12}\;
    ~Update $\theta_{\mathcal{R}_1}$ by $\theta_\mathcal{G}$ and $\theta_{\mathcal{R}_2}$ by $\theta_{\mathcal{F}}$ with EMA in Eq.\eqref{eq:7}\;
}
Return ${\theta}_\mathcal{G}$ and ${\theta}_\mathcal{F}$ 
\end{algorithm}

\section{Experiment}
\label{sec:exper}
In this section, we evaluate and analyze the proposed method through experiments. In section \uppercase\expandafter{\romannumeral5}-A, we start with the datasets and evaluation metrics. Then, section \uppercase\expandafter{\romannumeral5}-B presents the implementation details. Next, we compare our method with recent state-of-the-art methods on three benchmark datasets and verify the performance of our method in section \uppercase\expandafter{\romannumeral5}-C. Subsequently, section \uppercase\expandafter{\romannumeral5}-D conducts ablation studies to analyze the impact of different components. Furthermore, we visualize the qualitative results in section \uppercase\expandafter{\romannumeral5}-E.

\subsection{Datasets and Evaluation Metrics}
\noindent
\textbf{COCO Dataset}~\cite{10.1007/978-3-319-10602-1_48} 
is mainly divided into four subsets,~\emph{i.e.}, \emph{TRAIN}, \emph{VAL}, \emph{TEST-DEV} and \emph{TEST-CHALLENGE}.
It also contains 123K \emph{WILD} unlabeled images.
We randomly select 1K, 5K, and 10K from the \emph{TRAIN} to construct different training sets as the labeled set, and the remaining images in \emph{TRAIN} form the unlabeled set.
In addition, we conduct experiments where the entire \emph{TRAIN} serves as the labeled set and \emph{WILD} serves as the unlabeled set.

\noindent
\textbf{MPII Dataset}~\cite{6909866} involves 25K images and 40K human instance annotations, where the validation set contains 3K human instances.
\textbf{AI Challenger Dataset}~\cite{8785018} includes 210K images with 370K human instances.

\noindent
\textbf{Evaluation Metrics.} 
Following previous works~\cite{9710942,Huang_2023_CVPR},
we mainly report the commonly used metric of mAP~(mean AP over 10 OKS thresholds) on the COCO dataset to evaluate the performance and the metric of PCKh@0.5~\cite{andriluka20142d} on the MPII and AI Challenger datasets.

\subsection{Implementation Details}
We implement our method with PyTorch~\cite{paszke2019pytorch} and train the model using the Adam optimizer~\cite{2014Adam}.
Following the prior works~\cite{9710942,Huang_2023_CVPR}, we utilize the SimpleBaseline~\cite{xiao2018simple} for estimating heatmaps.
$\lambda$ is set as 0.5.
For the COCO dataset, the size of input images is resized to 256$\times$192, and the initial learning rate is set as 0.001, with the learning rate decaying by a factor of 10 at 70, 90 epoch in turn.
For the MPII and AI challenger datasets, input images are resized into 256$\times$256, the learning rate is 0.001. 
For a fair comparison, consistent with previous studies~\cite{9710942,Huang_2023_CVPR}, we adopt the same random rotation and random scale for easy data augmentation and hard data augmentation during training process.
Network $\mathcal{G}$ and Network $\mathcal{F}$ have close performance at the end, so we follow~\cite {9710942} to report their average accuracy.

\subsection{Comparison with State-of-the-Art Methods}
\noindent
\textbf{Results on COCO dataset}.
We compare our method with the state-of-the-art methods on the COCO dataset, and the experimental results are shown in Table~\ref{tab:2}. 
These methods contain fully supervised method (\emph{i.e.}, Supervised~\cite{xiao2018simple}) and semi-supervised methods (\emph{i.e.},  PseudoPose~\cite{yan2018spatial}, DataDistill~\cite{radosavovic2018data}, Single~\cite{9710942}, Dual~\cite{9710942} and SSPCM~\cite{Huang_2023_CVPR}).
The experiments are mainly conducted under 1K, 5K and 10K labeled data. 
From the results, we can see that our method outperforms the supervised training with only labeled data. 
Specifically, when ResNet18 is the backbone, our method surpasses the previous best performance by 4.0, 2.9, and 2.2 in terms of AP in 1K, 5K, and 10K, respectively. The reasons for the improved performance stem from: 1) As discussed previously, Dual and SSPCM utilize backpropagation to optimize networks, neglecting to retain crucial historical information. In contrast, our method utilizes the reviewer networks to retain important parameter information from both the teacher and student networks, thus bringing significant improvements; 2) Our method uncovers extra supervisory signals from both data and feature levels.
Compared to Dual and SSPCM, which only use the last stage of the backbone to estimate the heatmap, our method supervises the network by leveraging output feature from different stages. Moreover, our method utilizes the data augmentation strategy Keypoint-Mix to confuse keypoint features, thereby improving keypoints of discrimination.

In addition, following the previous works~\cite{9710942,Huang_2023_CVPR}, 
we compare with existing semi-supervised 2D HPE methods (\emph{i.e.}, Dual~\cite{9710942} and SSPCM~\cite{Huang_2023_CVPR}) under different backbone (\emph{i.e.}, ResNet50 and ResNet101) as shown in Table~\ref{tab:9}, where COCO \emph{TRAIN} severs as the labeled set and \emph{WILD} as the unlabeled set. It can be seen from the table that our method outperforms other methods in different backbone settings.

\begin{table}[t]
\caption{Comparison of AP results with state-of-the-art methods on COCO \emph{VAL} dataset, where 1K, 5K, and 10K samples from the COCO \emph{TRAIN} dataset are used as the labeled set and the remaining images in COCO \emph{TRAIN} as the unlabeled set. * denote the fully supervised method. The best results are highlighted in bold and the previous best record with underline.}
\label{tab:2}
\centering
\renewcommand{\arraystretch}{1.1}
\setlength{\tabcolsep}{1.1mm}{
\begin{tabular}{l|c|ccc}
\specialrule{0.12em}{1pt}{1pt}
\rowcolor{gray!40}Methods  & Backbone & 1K & 5K & 10K   \\ \hline
Supervised*   & ResNet18  & 31.5 & 46.4 & 51.1  \\ 
PseudoPose  & ResNet18   &  37.2     &  50.9    & 56.0 \\
DataDistill  & ResNet18   & 37.6 &  51.6  & 56.6   \\ 
Single  & ResNet18  &  42.1 &  52.3 &  57.3  \\
Dual & ResNet18  &44.6 & 55.6 &  59.6  \\
SSPCM  & ResNet18   &  \underline{46.9}  & \underline{57.5}  & \underline{60.7}   \\ \hline
\textbf{Ours}  & ResNet18  & \textbf{50.9}$\textcolor{blue}{\uparrow4.0}$     &  \textbf{60.4}$\textcolor{blue}{\uparrow2.9}$   &  \textbf{62.9}$\textcolor{blue}{\uparrow2.2}$     \\ \hline \hline
Supervised*  & ResNet50   & 34.4 &  50.3 &  56.3   \\ 
Dual  & ResNet50 & 48.7 & 61.2 & 65.0    \\
SSPCM  & ResNet50 &  \underline{49.4} &   \underline{61.6} &   \underline{65.4}    \\ \hline
  \textbf{Ours}   & ResNet50 &  \textbf{52.2} $\textcolor{blue}{\uparrow2.8}$     &  \textbf{63.5} $\textcolor{blue}{\uparrow1.9}$ &    \textbf{67.6} $\textcolor{blue}{\uparrow2.2}$    \\ 
\specialrule{0.12em}{1pt}{1pt}
\end{tabular}}
% \vspace{-1mm}
\end{table}

\begin{table}[t]
\caption{Comparison with existing methods on COCO \emph{VAL} dataset under different backbone, where COCO \emph{TRAIN} dataset is used as the labeled set and COCO \emph{WILD} dataset severs as the unlabeled set.  * denote the fully supervised method. The best results are highlighted in bold. }
\label{tab:9}
\centering
\renewcommand{\arraystretch}{1.2}
\setlength{\tabcolsep}{1.8mm}{
\begin{tabular}{l|c|cccc}
\specialrule{0.1em}{1pt}{1pt}
\rowcolor{gray!40}Methods  & Backbone & AP & AP .5 & AR & AR .5   \\ \hline
Supervised*   & ResNet50  & 70.9 &  91.4 & 74.2 & 92.3     \\ 
Dual  &  ResNet50 &  73.9 & 92.5 & 77.0 & 93.5  \\ 
SSPCM   & ResNet50     & 74.2 & 92.7 & 77.2 & 93.8     \\\hline
\textbf{Ours}   & ResNet50  & \textbf{74.9} & \textbf{93.5} & \textbf{77.6} & \textbf{94.0}\\ \hline \hline
Supervised*   & ResNet101 & 72.5 & 92.5 & 75.6 & 93.1 \\ 
Dual  & ResNet101&  75.3 & 93.6 & 78.2 & 94.1 \\ 
SSPCM  & ResNet101&  75.5 & \textbf{93.8} & 78.4 & 94.2 \\\hline
\textbf{Ours}   & ResNet101  & \textbf{75.8} & 93.6 &  \textbf{78.5}  & \textbf{94.3}  \\
\specialrule{0.1em}{1pt}{1pt}
\end{tabular}}
\end{table}

\begin{table}[t]
\caption{Comparison of PCKh@0.5 results with state-of-the-art methods Dual~\cite{9710942} and SSPCM~\cite{Huang_2023_CVPR} on MPII and AI Challenger datasets. ResNet18 is the backbone. The best results are highlighted in bold.}
\label{tab:3}
\centering
\renewcommand{\arraystretch}{1.2}
% \vspace{-2.2mm}
\setlength{\tabcolsep}{1.2mm}{
\begin{tabular}{l|cccccccc}
\specialrule{0.12em}{1pt}{1pt}
\rowcolor{gray!40} Methods & Hea  & Sho  & Elb & Wri & Hip & Kne & Ank  &Total  \\ \hline
Dual  & 95.6 & 93.8 & 85.0 &  78.4 & 85.8 & \textbf{79.4} & 74.2 & 85.3 \\ 
SSPCM  & 95.5 & 93.6 & 84.7 & 78.3 & 85.9 & \textbf{79.4}  & 74.3 & 85.3 \\  
\hline
\textbf{Ours}    &  \textbf{95.7} & \textbf{93.9} & \textbf{85.7} & \textbf{79.3} & \textbf{86.3} & 79.1 & \textbf{74.9} & \textbf{85.7} \\  
\specialrule{0.12em}{1pt}{1pt}
\end{tabular}}
\end{table}

\begin{table}[t]
\caption{Comparison of PCKh@0.5 results with state-of-the-art methods Dual~\cite{9710942} and SSPCM~\cite{Huang_2023_CVPR} on MPII dataset. All models utilize ResNet18 as the backbone. The best results are highlighted in bold.}
\label{tab:4}
\centering
% \vspace{-4mm}
\renewcommand{\arraystretch}{1.2}
\setlength{\tabcolsep}{1.2mm}{
\begin{tabular}{l|cccccccc}
\specialrule{0.12em}{1pt}{1pt}
\rowcolor{gray!40} Methods                  & Hea & Sho & Elb & Wri & Hip & Kne & Ank & Total \\ \hline
Dual  & 92.9 & 87.8 & 74.7 &  68.1 & 72.5 & 64.4 & 59.6 & 75.4      \\ 
SSPCM                 & 93.0 & 88.1 &  74.7 & 67.0 &  72.3 &  65.0 & 59.4 & 75.3\       \\ \hline
\textbf{Ours}                         &  \textbf{94.0} & \textbf{91.0} &\textbf{80.6} &\textbf{72.9} &\textbf{76.2} & \textbf{69.0} & \textbf{62.9} & \textbf{79.1}      \\   \specialrule{0.12em}{1pt}{1pt}
\end{tabular}}
\end{table}

\begin{table}[t]
\caption{The effects of different components on COCO dataset. MFL and KM denote Multi-level Feature Learning and Keypoint-Mix, respectively.
We report AP results under 
 different labeled data. The best results are highlighted in bold. }
 \label{tab:5}
\centering
\renewcommand{\arraystretch}{1.2}
\setlength{\tabcolsep}{3.3mm}{
\begin{tabular}{ccc|ccc}
\specialrule{0.12em}{1pt}{1pt}
\rowcolor{gray!40} MFL & KM  & Reviewer & 1K & 5K & 10K   \\ \hline
 $\times$ &  $\times$  &   $\checkmark$  & 45.7  & 56.9   & 60.6   \\ 
 $\checkmark$   &   $\times$   &  $\checkmark$  &  47.3 &  58.2    &  61.3   \\ 
  $\times$    &      $\checkmark$ &  $\checkmark$ & 48.4   &  58.6  &  61.8  \\   
  $\checkmark$  &  $\checkmark$  &    $\times$  & 44.8  & 56.8   & 60.5   \\ \hline
       $\checkmark$   &    $\checkmark$  &  $\checkmark$  & \textbf{50.9}      &   \textbf{60.4}   &   \textbf{62.9}    \\ 
\specialrule{0.12em}{1pt}{1pt}
\end{tabular}}
\end{table}

\begin{table}[t]
\caption{Impact of different feature learning stages on  COCO dataset with 1K labeled data. The best results are highlighted in bold. }
\label{tab:6}
\centering
\renewcommand{\arraystretch}{1.2}
\setlength{\tabcolsep}{4.0mm}{
\begin{tabular}{l|c|ccc}
\specialrule{0.12em}{1pt}{1pt}
\rowcolor{gray!40}  Methods & Backbone &  1  &  2  &   3    \\   \hline
\textbf{Ours}    & ResNet18  &  48.4  & \textbf{50.9} &  50.0  \\ 
\specialrule{0.12em}{1pt}{1pt}
 \end{tabular}}
\end{table}
 
\begin{table}[ht]
\caption{Impact of Keypoint-Mix with different number of keypoints on COCO dataset. We report the AP result under 1K labeled data. The best results are highlighted in bold. }
\label{tab:7}
\centering
\renewcommand{\arraystretch}{1.2}
\setlength{\tabcolsep}{2.7mm}{
\begin{tabular}{c|c|cccc}
\specialrule{0.12em}{1pt}{1pt}
\rowcolor{gray!40} Methods  & Backbone  & 3 & 5 & 7 & 9   \\ \hline
Our  &  ResNet18 &  50.2   & \textbf{50.9} & 50.0  &  49.8      \\  
\specialrule{0.12em}{1pt}{1pt}
\end{tabular}}
\end{table}

\begin{table}[t]
\caption{Ablation study of different data augmentation. We compared Keypoint-Mix (KM) with other methods such as   Cutout~\cite{devries2017improved},  
Mixup~\cite{zhang2017mixup},  CutMix~\cite{yun2019cutmix}, Rand Augment~\cite{cubuk2020randaugment}, JC~\cite{9710942} and SSCO~\cite{Huang_2023_CVPR}. 
Training is conducted on 1k labeled data from the COCO \emph{TRAIN} dataset, with testing on the COCO \emph{VAL} dataset. The best results are highlighted in bold.}
\label{tab:8}
\centering
\renewcommand{\arraystretch}{1.2}
\setlength{\tabcolsep}{4.0mm}{
\begin{tabular}{l|c|c|c}
\specialrule{0.12em}{1pt}{1pt}
\rowcolor{gray!40} Methods  &   Augmentation & Backbone   & AP   \\ \hline
Dual  &  JC   &  ResNet18   & 44.6   \\
SSPCM & SSCO  &  ResNet18   &46.9 \\  \hline
Ours &  Cutout  &  ResNet18   &  47.4 \\
Ours &  Mixup  &  ResNet18   &  46.2 \\
Ours &  CutMix  &  ResNet18   &  49.6 \\
Ours &  Rand Augment  &  ResNet18   &  49.5 \\
Ours &   JC  &  ResNet18   &  49.6  \\
Ours &   SSCO &  ResNet18   & 50.1 \\\hline
\textbf{Ours} &KM &  ResNet18    & \textbf{50.9} \\  
\specialrule{0.12em}{1pt}{1pt}
\end{tabular}}
\end{table}

\noindent
\textbf{Results on MPII and AI Challenger datasets}. 
We selected 10K samples from the MPII dataset as the labeled set and 100K samples from the AI-Challenger dataset as the unlabeled set for training, using the MPII validation for testing. The results are shown in Table~\ref{tab:3}, it can be observed that our method outperforms other semi-supervised methods Dual~\cite{9710942}, SSPCM~\cite{Huang_2023_CVPR} and proves the effectiveness of our method.

\noindent
\textbf{Results on MPII dataset}.
We conducted experiments using 1K samples as labeled data and 10K samples as unlabeled data in the MPII dataset, and the validation set of the MPII dataset is used to test. The results are presented in Table~\ref{tab:4}, and we can find that our method achieves optimal performance. This further demonstrates the generality of our method.

\subsection{Ablation Study}
\noindent
\textbf{Impact of different components}.
We evaluate the impact of different components of our method using labeled data with 1K, 5K, and 10K samples, as presented in Table~\ref{tab:5}.
We use ResNet18 as the backbone. 
First, we set up our model without Multi-level Feature Learning~(MFL) and Keypoint-Mix~(KM) as the baseline, and then gradually incorporate MFL and KM into the baseline.  
In addition, we perform an experiment in which the reviewer networks are removed, while MFL and KM are retained.
From the results, it can be observed that with the increase of different components, the results gradually improve under different labeled data used, which is sufficient to prove the effectiveness of different components.

\begin{figure*}[t]
\begin{center}
\includegraphics[width=1\linewidth]{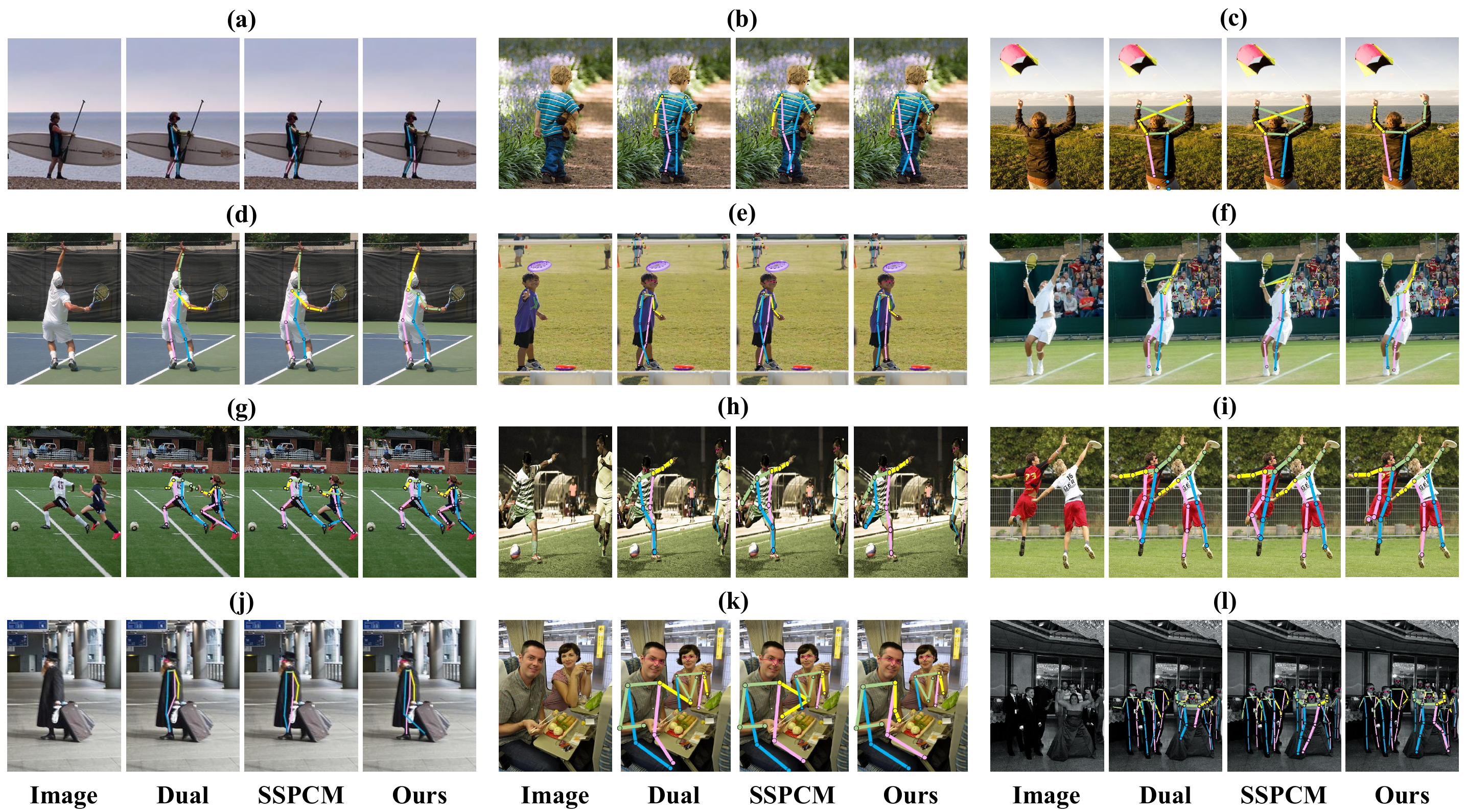}
    \end{center}
    \vspace{-2mm}
    \caption{Qualitative comparison of our method and other semi-supervised 2D HPE methods Dual~\cite{9710942} and SSPCM~\cite{Huang_2023_CVPR} on COCO \emph{VAL} dataset, where 
    all models are trained with 1K labeled data using ResNet18 as the backbone. 
    The first and second rows indicate single-person scenario, the third row denotes multiple-person scenario, and the fourth row represents occlusion scenario.}
\label{fig:4}
\end{figure*}

\begin{table}[t]
\caption{The effects of employing different network structures for Teacher and Student with state-of-the-art methods Dual~\cite{9710942} and SSPCM~\cite{Huang_2023_CVPR} on COCO \emph{VAL} dataset. The best results are highlighted in bold.  * denote the fully supervised method.}
\label{tab:10}
\centering
\renewcommand{\arraystretch}{1.2}
\setlength{\tabcolsep}{2.3mm}{
%\vspace{-1mm}
\begin{tabular}{l|cc|cc}
\specialrule{0.1em}{1pt}{1pt}
\rowcolor{gray!40}  Methods & Teacher & Student   & 5K & 10K   \\ \hline 
Supervised*   & - & ResNet18    & 46.4 & 51.1 \\ 
Supervised*  & - & ResNet50     &  50.3 &  56.3 \\ \hline\hline
Dual  &  ResNet18 &  ResNet18  &  55.6 &  59.6 \\
Dual  &  ResNet50 & ResNet50 & 61.2 & 65.0 \\ 
Dual &  ResNet50 & ResNet18   & 57.2 & 60.4  \\\hline \hline
SSPCM  &  ResNet18 &  ResNet18   &  57.5 &  60.7 \\
SSPCM  &  ResNet50 &  ResNet50   &  61.6 &  65.4 \\
SSPCM  &  ResNet50 &  ResNet18   &  58.9 &  61.9 \\\hline \hline
\textbf{Ours} &  ResNet18 &  ResNet18         &   \textbf{60.4}   &  \textbf{62.9}   \\
\textbf{Ours}  &  ResNet50 &  ResNet50    &  \textbf{63.5}  &    \textbf{67.6}  \\  
\textbf{Ours}  &  ResNet50 &  ResNet18    &  \textbf{61.4}  &    \textbf{63.7}  \\ 
\specialrule{0.1em}{1pt}{1pt}
\end{tabular}}
\end{table}

\begin{figure}[t]
\begin{center}
     \includegraphics[width=1\linewidth]{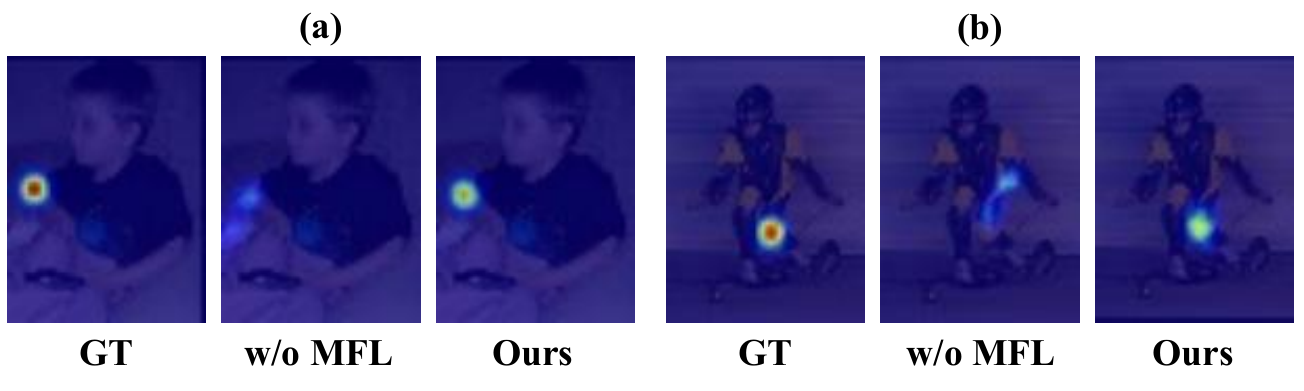}
\end{center}
\vspace{-3mm}
\caption{Heatmap visualization of two samples from COCO dataset. 
The columns are arranged from left to right as follows: ground truth (GT), heatmap estimation results of our method without using the Multi-level Feature Learning (w/o MFL), and heatmap estimation results of our full method.}
\label{fig:5}
\end{figure}

\begin{figure}[t]
\begin{center}
\includegraphics[width=1.0\linewidth]{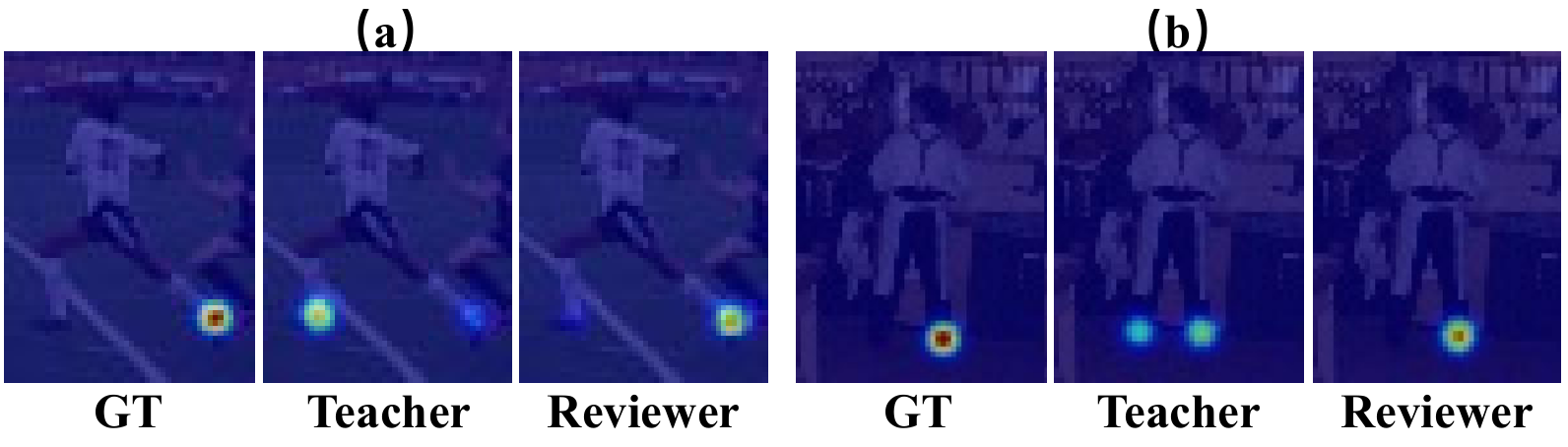}
    \end{center}
        \vspace{-3mm}
    \caption{Heatmap visualization of two samples on COCO dataset. Arranged from left to right as follows: ground truth (GT), heatmap estimation from the teacher network, and heatmap estimations from the reviewer network.}
\label{fig:6}
\end{figure}

\noindent
\textbf{Impact of Multi-level Feature Learning.}
We perform experiments on different feature learning stages using the COCO dataset under 1k labeled samples in Table~\ref{tab:6}, where ``1" denotes the final stage, ``2" and ``3" represent the last two and three stages, respectively.  
The results demonstrate that our model achieves the best performance when the last two stages are used to learn the relationship between keypoints.

\noindent
\textbf{Impact of Keypoint-Mix strategy.}
We explore the effect of Keypoint-Mix (KM) with different numbers of keypoints trained on 1k labeled data of COCO in Table~\ref{tab:7}. We find that our method achieves optimal performance when the number of keypoints ($K$) is 5.
Meanwhile, the results indicate that the performance of the KM strategy varies less with different numbers of keypoints, demonstrating its robustness and insensitivity to hyperparameter variations. 
In addition, to further prove the effectiveness of the KM, we compare it with other data augmentation methods, \emph{i.e.}, Cutout~\cite{devries2017improved}, Mixup~\cite{zhang2017mixup},  CutMix~\cite{yun2019cutmix}, Rand Augment~\cite{cubuk2020randaugment}, JC~\cite{9710942} and SSCO~\cite{Huang_2023_CVPR}), the results shown in Tabel~\ref{tab:8}. 
The results indicate that our KM outperforms other methods. 
We argue that the validity of KM lies in its ability to mix features from different keypoints, thus blurring the boundaries between keypoint features and making it difficult to distinguish between confused keypoint features, hence further motivating the network to improve the discernable ability for keypoints.

\noindent
\textbf{Impact of different network structures}.
We evaluate the impact of different network structures for teacher and student in 5K and 10K labeled data and compare with other methods
(\emph{i.e.}, Supervised~\cite{xiao2018simple},  Dual~\cite{9710942} and SSPCM~\cite{Huang_2023_CVPR}), the results are shown in Table~\ref{tab:10}. 
In the table, our method achieves competitive performance when network $\mathcal{G}$ and network $\mathcal{F}$ using different backbones and different labeled data.  
In addition, it can also be seen that when network $\mathcal{G}$ exploits ResNet50 and network $\mathcal{F}$ adopts ResNet18, the performance improves compared to using ResNet18 in both networks $\mathcal{F}$ and $\mathcal{G}$.
This improvement stems from the accurate supervision provided by ResNet50 over ResNet18, resulting in enhanced estimation performance. The results demonstrate that our method enables lightweight and large models to learn together and improve the accuracy of estimation compared to existing methods.

\subsection{Qualitative Results}
To further help understand the effect of our method, we present some qualitative results in this subsection. 
First, we present qualitative results of different examples from the COCO \emph{VAL} dataset in Figure~\ref{fig:4}. All models train on 1k labeled data with ResNet18 serving as the backbone. 
We show (a)-(f) for the single-person scenario in the first and second rows, (g)-(i) for the multi-person scenario in the third row, and (j)-(l) for the occlusion scenario in the fourth row, respectively. As can be seen from the figure, our method outperforms Dual~\cite{9710942} and SSPCM~\cite{Huang_2023_CVPR} in various scenarios.

For example, in the single-person scenario, our method predicts the keypoint positions in the lower body better than Dual and SSPCM in (a) and (b). In addition, 
we can accurately predict the wrist, elbow, and shoulder positions on both sides compared to Dual and SSPCM in (d) and (f). Meanwhile, our method can better predict the correct position of the children's ankles in (e), unlike Dual and SSPCM, which incorrectly swap the left and right ankle positions. 
As evident in multi-person scenario, our method accurately predicts the leg posture of the right person (g) compared to Dual and SSPCM, respectively.
Also, we can accurately predict the left person's knee position in (h) and the left person's ankle position in (i).
Additionally, our method also performs better in occlusion scenario, accurately predicting the ankle in the case of suitcase occlusion (j) and the arm position of the man near the woman's side (k), as well as the posture of the woman on the left side and the person behind her (l). This implies that our method can better learn the relationships between keypoints, thereby facilitating improved network learning and leading to accurate prediction.

In addition, we present the heatmap results of our method, both with and without the Multi-level Feature Learning~(MFL) strategy, in Figure~\ref{fig:5}. All models train on 1k labeled data with ResNet18 serving as the backbone. 
From the figure, our method inaccurately predicts the keypoint locations when without employing the MFL strategy. 
More precisely, our method without MFL only provides rough estimations of the elbow position in (a) and rough knee position estimates in (b). 
In contrast, our method accurately predicts the position in two samples. 
The reason is that we utilize additional spatial information as a supervisory signal, assisting in precisely localizing keypoints.

Then, we show the different heatmap results of the teacher and reviewer in Figure~\ref{fig:6}. 
As shown in (a), when the teacher network incorrectly predicts the right ankle as the left ankle, the reviewer network predicts the right ankle correctly. Similarly, when the teacher network predicts the ankle with lower confidence, the reviewer network offers a more precise location in (b).
The above results show that the information from the reviewer network and teacher network is complementary yet distinct.
Therefore, we argue that the reviewer network can provide diverse feedback to the student network, thereby enhancing robustness.

\section{Conclusion}
\label{sec:con}
In this paper, we present a novel \emph{Teacher-Reviewer-Student} framework for the semi-supervised 2D human pose estimation task, where the teacher network is used to predict results for unlabeled data to guide the student network's training, and reviewer networks are proposed to store important historical parameters training information while providing additional supervision. 
In addition, we introduce a Multi-level Feature Learning strategy to enrich the supervisory signals by utilizing different features, and a new data augmentation named Keypoint-Mix to perturb the pose information while retaining crucial pose details. Comprehensive experiment results demonstrate the effectiveness and superiority of our proposed method on public benchmarks. In the future, we plan to integrate semi-supervised pose estimation with tasks such as anomaly action detection, and design more general and efficient models through multi-task learning applied in the social security governance.

\end{document}